\documentclass[10pt,twocolumn,letterpaper]{article}

\usepackage{wacv}
\usepackage{times}
\usepackage{epsfig}
\usepackage{graphicx}
\usepackage{amsmath}
\usepackage{amssymb}
\usepackage{subcaption}

\usepackage{authblk}

\def\wacvPaperID{594} % Enter the WACV Paper ID here

\wacvfinalcopy % *** Uncomment this line for the final submission

\ifwacvfinal
\fi

\ifwacvfinal
\usepackage[breaklinks=true,bookmarks=false]{hyperref}
\else
\usepackage[pagebackref=true,breaklinks=true,colorlinks,bookmarks=false]{hyperref}
\fi

\begin{document}

%%%%%%%%% TITLE
\title{Improving Apparel Detection with Category Grouping and Multi-grained Branches}

\author[1]{Qing Tian\thanks{The work was done during Tian's internship at Amazon as an applied scientist.}}
\author[2]{Sampath Chanda}
\author[2]{Amit Kumar K C}
\author[2]{Douglas Gray}
\affil[1]{ECE Department, McGill University, Montreal, QC, Canada}
\affil[2]{Amazon Visual Search \& AR, Palo Alto, CA, USA}

\maketitle
%\thispagestyle{empty}

%%%%%%%%% ABSTRACT
\begin{abstract}
Training an accurate object detector is expensive and time-consuming. One main reason lies in the laborious labeling process, \textit{i.e.}, annotating category and bounding box information for all instances in every image. In this paper, we examine ways to improve performance of deep object detectors without extra labeling. 
We first explore to group existing categories of high visual and semantic similarities together as one super category (or, a superclass). Then, we study how this knowledge of hierarchical categories can be exploited to better detect object using multi-grained RCNN top branches\footnote{The term \emph{multi-grained} is used to describe features of different abstractness and granularity levels, corresponding to \emph{coarse-} and \emph{fine-}grained categories. RCNN: Regional Convolutional Neural Networks \cite{ren2015faster}.}. Experimental results on DeepFashion2 and OpenImagesV4-Clothing reveal that the proposed detection heads with multi-grained branches can boost the overall performance by 2.3 mAP for DeepFashion2 and 2.5 mAP for OpenImagesV4-Clothing with no additional time-consuming annotations. More importantly, classes that have fewer training samples tend to benefit more from the proposed multi-grained heads with superclass grouping. In particular, we improve the mAP for last 30\% categories (in terms of training sample number) by 2.6 and 4.6 for DeepFashion2 and OpenImagesV4-Clothing, respectively.
\end{abstract}

%%%%%%%%% BODY TEXT
\section{Introduction}

In the deep learning era, accurate object detection relies on a large amount of training data and annotations. However, instance-level annotations, such as bounding boxes, are expensive to collect compared to image-level labels. Therefore, it is important to make full use of currently available data and annotations. To this end, we propose to improve detection performance using multi-grained detection heads with superclass grouping in this paper. We integrate and utilize information from different abstraction levels during both training and inference. Such detectors that can extract and utilize hierarchical features are desirable in many real-world scenarios. For example, in case of fashion items, the features that separate different kinds of tee-shirts should be on different levels with those that distinguish tee-shirts from skirts. In practice, most items in catalogs of online retailers are organized hierarchically so that data and ground-truth labels are usually collected and used for training in a multi-grained fashion. However, most existing object detection algorithms such as SSD~\cite{liu2016ssd}, YOLO and its variants~\cite{redmon2016you,redmon2017yolo9000,redmon2018yolov3}, Mask RCNN~\cite{he2017mask} are designed to exploit labels in a flat or uniform manner. Consequently, these models cannot benefit from hierarchical nature of the data.

In addition, most real-world datasets are imbalanced in that some categories have more samples than others. Strategies such as data resampling~\cite{he2009learning, chawla2002smote} and cost-sensitive learning~\cite{krawczyk2014cost, ting2000comparative, zhou2010multi} can be adopted to alleviate this imbalance problem. Relatively speaking, fewer deep learning approaches~\cite{zhou2005training,wang2016training, khan2017cost,wang2017learning} have been proposed to address this issue for object detection. Detectors trained on imbalanced datasets can lead to low detection accuracy for small classes (\emph{i.e.,} classes with a limited number of samples). We hypothesize that closely-related fine-grained categories may utilize similar coarse-grained features. We attempt to improve performance on small categories via learning shared coarse features with the help of data from related large categories. In this way, we make full use of available bounding box annotations.

%The remainder of the paper is organized as follows: the relevant literature is reviewed in Sec.~\ref{sec:literature}. In Sec.~\ref{sec:multistageheads}, our proposed category grouping and detection heads with multi-grained branches are introduced. Sec.~\ref{sec:experiments} describes our experimental results and compares our proposed frameworks to the baseline in terms of overall and per-category AP scores. Sec.~\ref{sec:futurework} discusses future work and Sec.~\ref{sec:conclusion} concludes the paper.

%-------------------------------------------------------------------------

\section{Related works}\label{sec:literature}

In this section, we review several deep learning based object detection approaches. One of the dominant paradigms in modern object detection is based on a two-stage approach. In the first stage, a set of candidate proposals are generated using techniques like Selective Search~\cite{uijlings2013selective} and EdgeBoxes~\cite{zitnick2014edge}. The second stage refines the proposals into a final set of bounding boxes along with their corresponding categories. RCNN~\cite{girshick2014rich} and its variants, namely Fast RCNN~\cite{girshick2015fast} and Faster RCNN~\cite{ren2015faster}, belong to this category of detectors. Mask RCNN~\cite{he2017mask} is based on Faster RCNN with a separate mask branch added for instance segmentation. In an effort to produce more accurate bounding boxes, multi-step detectors have also been proposed that explore different ways of gradually refining the detections. Some examples are multi-region detector~\cite{gidaris2015object}, CRAFT~\cite{yang2016craft}, AttractioNet~\cite{gidaris2016attend}, Cascade R-CNN ~\cite{cai2018cascade}.

Compared to the two-stage paradigm, one-stage object detectors have gained popularity mainly due to their efficiency. OverFeat~\cite{sermanet2013overfeat} is one of the first one-stage object detectors. More recently, SSD~\cite{liu2016ssd} and YOLO~\cite{redmon2016you} (including its variants such as YOLOv3~\cite{redmon2018yolov3}) have reduced the detection accuracy gap between one-stage and two-stage detectors while improving on the computation efficiency. To achieve better accuracy without much extra computation, Lin~\etal~\cite{lin2017focal} have proposed \emph{focal loss} to assign higher weights to hard examples dynamically. Another variant of YOLO, called YOLO9000 ~\cite{redmon2017yolo9000}, detects over 9000 object categories by taking advantage of both the abundance of fine-grained category labels in ImageNet~\cite{imagenet_cvpr09} and the plentifulness of coarse-class bounding box annotations in the MS-COCO dataset~\cite{lin2014microsoft}. Another work that uses coarse- and fine-grained category organization is~\cite{yang2019detecting}. It detects at the level of fine grained classes, while only requiring a small set of bounding box annotations at coarse-grained class level.

In most modern detectors such as the above-mentioned Faster RCNN \cite{ren2015faster}, the same level of (top) features is used to detect and classify different categories. However, representations suitable for different categories may lie on different abstraction levels. Previous works, such as skip connection~\cite{he2016deep,huang2017densely} and layer aggregation~\cite{yu2018deep}, have been shown to be effective for integrating different-level information. As far as we know, few, if any, works have experimented with utilizing multi-level information in distinct detection heads (after ROI determination). Also, none has attempted to inject multi-level supervision information on such heads to extract features at different granularities.
Our work builds on the two-stage Faster RCNN backbone and we propose detection heads with multi-grained branches to exploit information from different abstractness levels. Unlike Mask RCNN~\cite{he2017mask} (another Faster-RCNN multiple-branch extension as mentioned previously), our multiple branches capture multi-grained information and tackle the same final task (fine-grained apparel detection in our case). To train these multi-grained branches, we first group the categories into hierarchical manner (\emph{i.e.,} superclass grouping).

%------------------------------------------------------------------------
\section{Category grouping and multi-grained detection heads}
\label{sec:multistageheads}

In this section, we first discuss category grouping strategies and then present different detection head architectures to improve the overall detection performance as well as the performance for minority categories.

\subsection{Superclass grouping for multi-grained detection heads}
\label{section:superclass_grouping}
\begin{figure}[h]
\begin{center}
\includegraphics[width=\linewidth, height=1.0in]{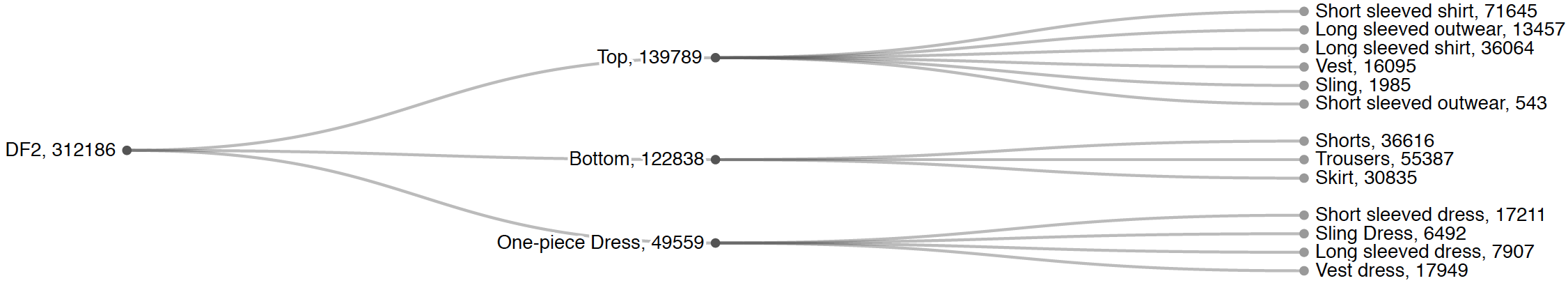}
\end{center}
\caption{Category grouping and instance number in each category of DeepFashion2 training set.}
\label{fig:deepfashion2tree}
\end{figure}

First, we need to build a hierarchical structure of the data. If the dataset contains hierarchical information, we respect the original hierarchy or superclass grouping. If not, the WordNet~\cite{miller1990introduction} hierarchy can be used as a guidance (as many works do~\cite{redmon2017yolo9000,deng2012hedging}). In this paper, we group fine-grained categories into coarse-grained super-categories based on visual and functional similarity. 
It is worth noting that the extra superclass grouping/labeling is very cheap. It simply becomes a dictionary checkup procedure after a hierarchy containing supercategory and category pairs is defined. This cost can be neglectible in most cases.
To be more specific, we use two publicly available datasets, namely DeepFashion2~\cite{ge2019deepfashion2} and OpenImagesV4~\cite{kuznetsova2018open} in this paper. We group the categories as follows:

\textbf{DeepFashion2} contains 491K images of 13 clothing categories from both commercial shopping stores and consumers, out of which 312186 training instances and 52490 validation instances are publicly available. For our detection task, only category and bounding box information are utilized in our experiments. The 13 categories are grouped into 3 super categories, namely top, bottom, and one-piece dress, as shown in Figure~\ref{fig:deepfashion2tree}.

\begin{figure}[h]
\begin{center}
\includegraphics[width=\linewidth, height=3.25in]{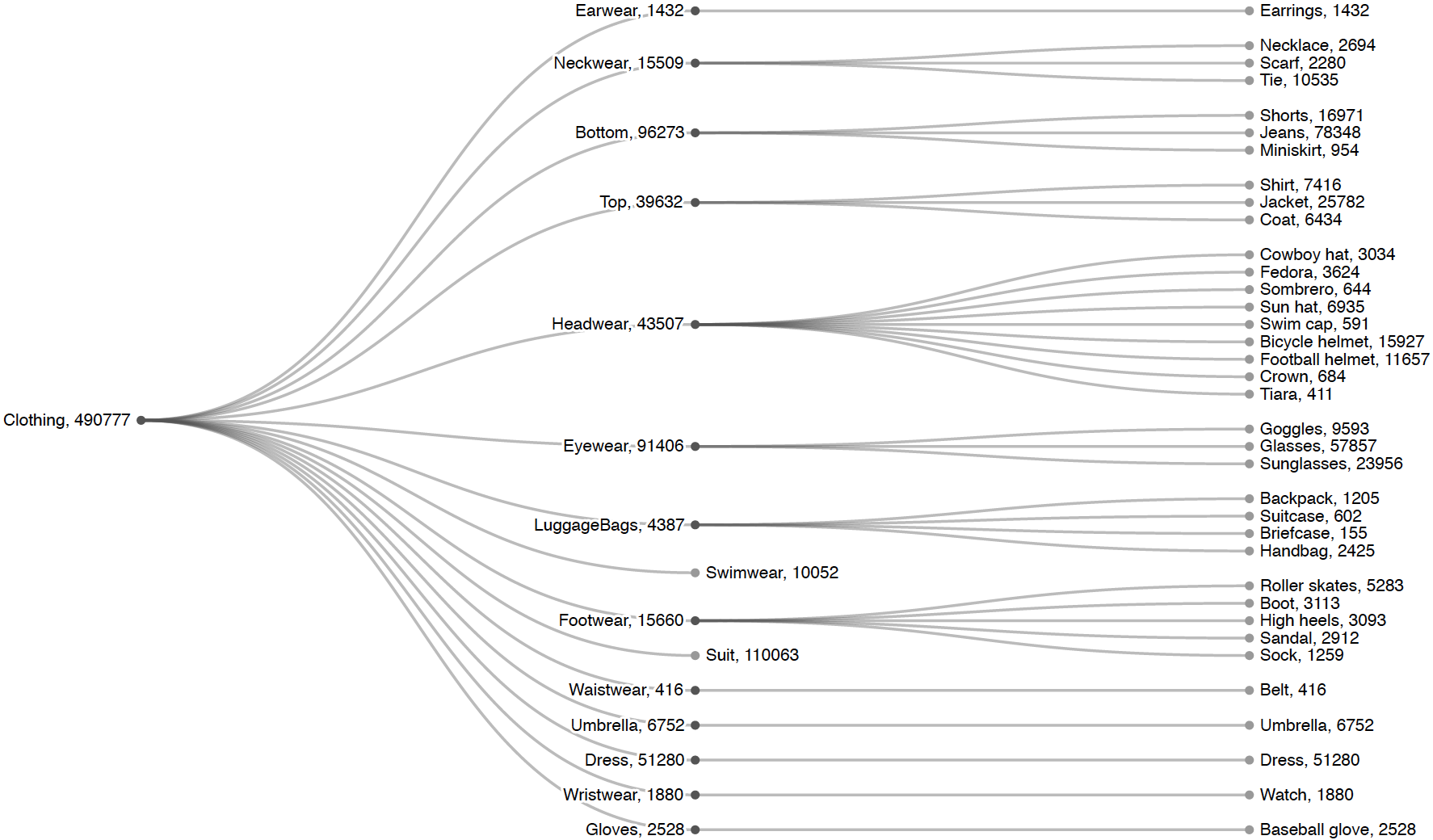}
\end{center}
\caption{Category grouping and instance number in each category of OpenImagesV4-Clothing training set. }
\label{fig:openimagestree}
\end{figure}

\textbf{OpenImagesV4}~\cite{kuznetsova2018open} contains 15.4 million bounding boxes for 600 categories on 1.9 million images, rendering it the largest existing public dataset with object location annotations (approximately 15 times more bounding boxes than ImageNet~\cite{imagenet_cvpr09} and MS-COCO~\cite{lin2014microsoft}). For our purpose, we select the subset of clothing instances, which we refer to as OpenImagesV4-Clothing dataset in the following.  There are 490,777 bounding box instances for training and 3,897 instances for validation in total in OpenImagesV4-Clothing dataset. While grouping the categories, we respect the hierarchy provided by the dataset\footnote{\url{https://storage.googleapis.com/openimages/2018_04/bbox_labels_600_hierarchy_visualizer/circle.html}} with a few changes. In particular, we perform the following modifications (A $\rightarrow$ B indicates class B of superclass A in the original dataset):
\begin{itemize}
	\item {We merge \emph{helmet}, \emph{hat}, and \emph{fashion accessory} $\rightarrow$ \emph{hat} as \emph{headwear}.}
	\item {We merge \emph{fashion accessory} $\rightarrow$ \emph{necklace, tie}, and \emph{clothing} $\rightarrow$ \emph{scarf} as \emph{neckwear}.}
	\item {We group \emph{clothing} $\rightarrow$ \emph{shorts}, \emph{trousers} $\rightarrow$ \emph{jeans}, and \emph{skirt} $\rightarrow$ \emph{miniskirt} together as bottom. }
\end{itemize}
Figure~\ref{fig:openimagestree} demonstrates the super category and category relationship with the number of training instances added to each category. 

In the following section, we describe how we propose to exploit the resulting hierarchical data structure by designing coarse- as well as fine-grained classifiers on top of Faster RCNN.

\subsection{Detection heads with multi-grained branches}
\begin{figure*}[!ht]
\centering
\begin{subfigure}{0.45\linewidth}
    \includegraphics[height=1.55in]{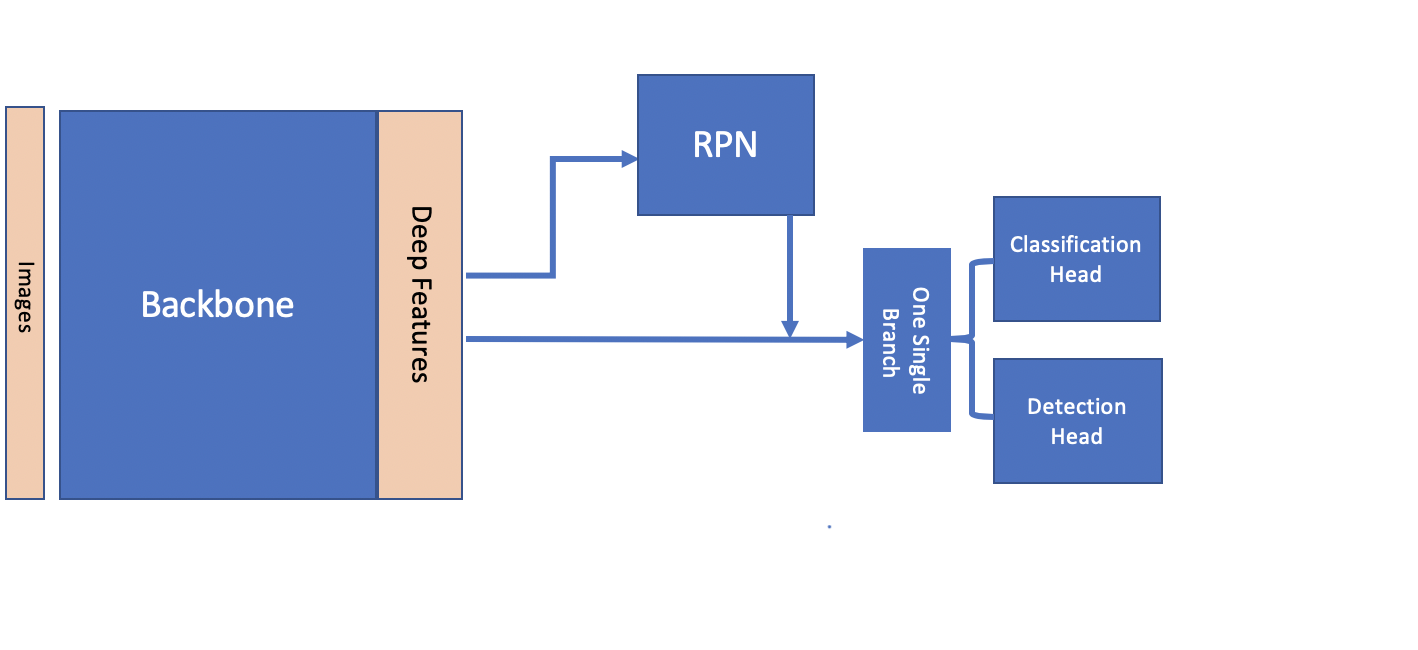}
    \caption{Single detection head in Faster RCNN}
    \label{fig:single}
\end{subfigure}
~
\begin{subfigure}{0.45\linewidth}
    \includegraphics[height=1.55in]{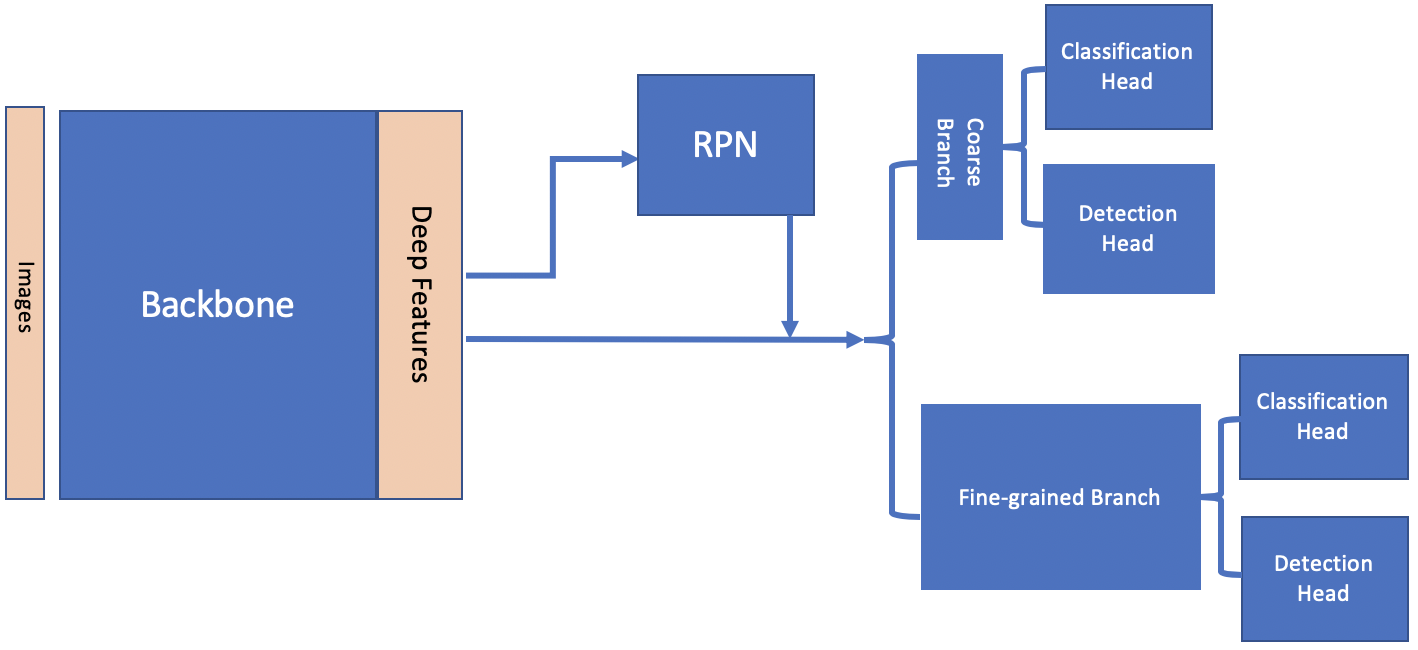}
    \caption{Multi-grained detection heads with Faster RCNN backends}
    \label{fig:twostage}
\end{subfigure}
\\\vspace{0.25in}
%\begin{subfigure}{0.45\linewidth}
%   \includegraphics[width=\linewidth, height=1.4in]{twostage.png}
%   \caption{RCNN with double-stage detection heads (with grouping)}
%     \label{fig:twostagegrouping}
%\end{subfigure}
%
\begin{subfigure}{\linewidth}
    \begin{center}
    \includegraphics[height=1.75in]{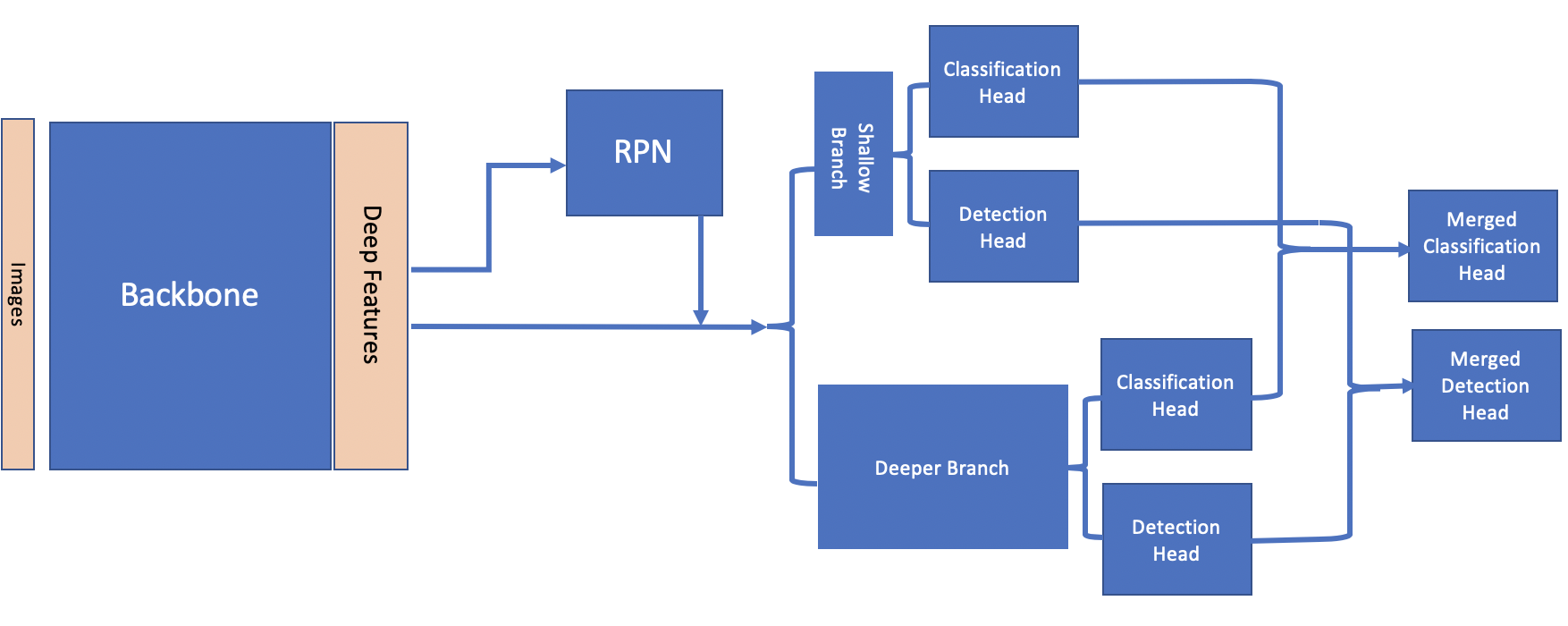}
    \caption{Merged detection heads with Faster RCNN backends}
    \label{fig:mergehead}
    \end{center}
\end{subfigure}
\caption{Detection heads with single or multi-grained branches on Faster RCNN backends. RPN: region proposal network. Blue blocks represent network structures while light orange blocks indicate feature maps and input (only important feature maps are shown). In our experiments, more layers are added to the one single branch in (a) to match the size of others.}
\label{fig:multistageheads}
\end{figure*}

In Faster RCNN, after backbone feature extraction and region-of-interest (ROI) proposal stages, a single branch with one set of bounding box regression and classification heads are plugged on top of the pooled/aligned ROI features. Consequently, it cannot integrate information from different abstract levels explicitly or utilize labels in a hierarchical manner. Instead, we propose to add several branches of different depths to the ROI features. Different depths of feature extraction produce features of different granularities and abstractness levels, on which our localization and classification are based. We refer such head structures as \emph{multi-grained} heads or detection heads with \emph{multi-grained} branches in this paper. Each branch can be any type of networks that help increase feature abstractness, \textit{e.g.}, convolutional or fully connected. The multi-grained branch ideas explored in this paper are demonstrated in Figure~\ref{fig:multistageheads}. In our datasets, the labels have been organized in two levels. Therefore, we only study detection heads of two granularity levels here. They can be easily extended to cases with more granularity levels.

Figure~\ref{fig:single} shows the original Faster RCNN structure~\cite{ren2015faster}. We choose ROI alignment \cite{he2017mask} instead of ROI pooling \cite{ren2015faster} for all our configurations. 

In Figure~\ref{fig:twostage}, two detection branches of different depths are added to the ROI-aligned feature maps. With more depths, the longer branch is expected to learn more abstract and finer-grained features than the shallower branch. The same number of neurons are on top of both the coarse- and fine-grained branches. The number is equal to the number of total fine-grained categories. Fine-grained supervision is applied on the fine-grained branch. We can inject either superclass (with grouping) or fine-grained class (w/o grouping) information onto the coarse branch. The latter is equivalent to grouping each category as a separate super category on its own. For the `with grouping' option, the amount of work needed to perform category-level grouping is negligible compared to instance-level annotation. We use only the fine-grained bounding box regression head to evaluate localization performance.

In addition, based on Figure~\ref{fig:twostage}, we add extra merging layers that combine information from coarse and fine grained stages as shown in Figure~\ref{fig:mergehead}. Compared to other parts of the network (backbone, RPN, and head branches), a merging layer is light. It only consists of one fully connected layer to integrate the outputs from different abstract levels and produce one final set of results. The lower layers to be merged are first concatenated before being fully connected with the merging layer. One difference of Fig.~\ref{fig:mergehead} from Fig.~\ref{fig:twostage} is that apart from the supervision added at the end, no separate coarse-level supervision is applied.

%\subsection{Superclass grouping on multi-grained detection heads}
%Based on Figure~\ref{fig:twostage}, we inject super class grouping knowledge into the coarse branch. Super categories are used here in contrast to fine-grained categories in the other branch. For datasets with no or partial hierarchy information available already, manual grouping is performed to put together finer categories with visual and semantic similarity. The amount of work needed to perform category-level grouping is negligible compared to instance-level annotation.

\textbf{Loss function:} For Figure~\ref{fig:single} and~\ref{fig:mergehead}, standard softmax cross entropy and smooth $L_1$ loss~\cite{girshick2015fast} are used as loss functions for box head classification and regression, respectively. For multi-grained branches without the additional merging head (Figure~\ref{fig:twostage}), we modify these losses to:
\begin{equation}
\label{eq:multistageentropy}
E_{\mbox{\tiny ce}} = -\sum_{l=1}^L \sum_{c=1}^C w_k^{(l)} t_c^{(l)} \ln y_c^{(l)}
\end{equation}
and
\begin{equation}
\label{eq:multistageloc}
E_{\mbox{\tiny loc}} = \sum_{l=1}^L \sum_{i \in (x,y,w,h)} w_r^{(l)} * smooth_{L_1}(o_i^u - v_i)
\end{equation}

\noindent where $E_{ce}$ and $E_{loc}$ correspond to the cross entropy and bounding box regression losses, $l$ indicates level or stage, $L$ is the number of levels used ($L=2$ in our case), $c$ represents category (including background), $y$ is the set of predicted results after softmax, t is the set of ground-truth labels (one-hot encoded in our case), $w_k$ and $w_r$ are stage-weighting factors for classification and bounding box regression. $smooth_{L_1}$ stands for the smooth $L_1$ function as proposed in~\cite{girshick2015fast}. $o^u$ is predicted bounding-box offsets tuple for the true class $u$ and $v$ is ground-truth bounding-box regression target. In our implementation, $L=2$ and $w_r$ for the coarse branch is 0. We drop the loss component from coarse branch to avoid the complexity of bounding box merging logic from multiple branches.  The total loss of the framework is the sum of the above two losses together with the binary classification and localization regression losses of the RPN subnetwork.

%------------------------------------------------------------------------
\section{Experiments}\label{sec:experiments}

In this section, we present the implementation details (Sec. \ref{sec:implementdetails}) and the evaluation metrics (Sec. \ref{sec:evalmetrics}) for our proposed solution. Finally, we present our quantitative and qualitative results and compare the proposed approaches and the baseline (Sec. \ref{sec:expresults}).

\subsection{Implementation details}
\label{sec:implementdetails}
Our implementations are based on Detectron2~\cite{wu2019detectron2}. Modifications are made to support detection heads with multi-grained branches and superclass grouping. Our frameworks are trained with a minibatch size of 24 for 200k iterations using a stochastic gradient descent (SGD) optimizer. The base learning rate is 0.005 and is decreased twice at 150K and 175K iterations by a factor of 10. We use a weight decay of 0.0001 and momentum of 0.9. For each RPN anchor, 5 scales and 3 aspect ratios are used to generate original anchor boxes. An anchor box is considered positive if its IoU with a groundtruth box exceeds 0.5. All the frameworks studied adopt a ResNet101 backbone~\cite{he2016deep} with Feature Pyramid Network (FPN)~\cite{lin2017feature} support. The shared backbone is pre-trained on the MS-COCO dataset~\cite{lin2014microsoft}. The coarse-grained and fine-grained branches consist of one and two fully connected layers of 1024 neurons, respectively. Parameter size differences between multi-grained detection heads with and without merged heads are negligible. As for the baseline model, we add more layers to the single branch after ROI alignment in order to match the size of our multi-grained approach. In our experiments, all competing frameworks have approximately 73M trainable parameters.

\subsection{Metrics}
\label{sec:evalmetrics}
We adopt mean average precision (mAP) to evaluate our methods. In our context, the overall mAP score is calculated by taking the mean AP over all classes and over all IoU thresholds.  Following the standard COCO metrics, for each framework on each dataset, we report AP (averaged over IoU thresholds), AP50 (IoU threshold 0.5), AP75 (IoU threshold 0.75), and AP$_s$ (small objects, area $< 32^2$), AP$_m$ (medium size objects), and AP$_l$ (large objects, area $> 96^2$).

\subsection{Results}
\label{sec:expresults}
In this section, we first present our quantitative results. Our focuses are on both overall mAP and Per-category AP (especially for small categories). Then, qualitatively, we show some success and failure cases of our approaches.
\subsubsection{Overall mAP}
Table~\ref{tab:deepfashion2results} and~\ref{tab:oiv4clothingresults} show the detection results on DeepFashion2 and OpenImagesV4-Clothing, respectively.

\begin{table*}
\begin{center}
\begin{tabular}{|l|c|c|c|c|c|c|}
\hline
 & $AP$ & $AP50$ & $AP75$ & $AP_s$ & $AP_m$ & $AP_l$ \\
\hline\hline
Baseline & 67.02 & 79.90 & 75.61 & 40.05 & 47.09 & 67.23 \\
W/O Group & 67.54 & 79.87 & 75.88 & 25.10 & 47.74 & 67.79\\
Merged & 68.80 & 81.55 & 77.40 & 31.75 & 50.80 & 69.04 \\
W Group & 69.02 & 81.60 & 78.12 & 31.75 & 51.16 & 69.23 \\
\hline
\end{tabular}
\end{center}
\caption{Results on the validation set of DeepFashion2. ‘Baseline’ stands for the baseline Faster RCNN (Fig.~\ref{fig:single}), ‘W/O Group’ indicates the naive multi-grained branch approach without grouping (Fig.~\ref{fig:twostage}), ‘W Group’ refers to the multi-grained branch with superclass grouping (Fig.~\ref{fig:twostage}), ‘Merged’ represents the merged heads approach (Fig.~\ref{fig:mergehead}).}
\label{tab:deepfashion2results}
\end{table*}

\begin{table*}
\begin{center}
\begin{tabular}{|l|c|c|c|c|c|c|}
\hline
 & $AP$ & $AP50$ & $AP75$ & $AP_s$ & $AP_m$ & $AP_l$ \\
\hline\hline
Baseline &  47.88 & 67.19 & 52.45 & 8.21 & 29.05 & 52.21 \\
W/O Group & 48.40 & 67.79 & 53.53 & 8.96 & 31.31 & 52.85\\
Merged & 49.48 & 68.59 & 54.98 & 8.79 & 29.23 & 54.12 \\
W Group & 49.50 & 68.34 & 55.12 & 9.77 & 29.19 & 53.91 \\
\hline
\end{tabular}
\end{center}
\caption{Results on the validation set of OpenImagesV4-Clothing. ‘Baseline’ stands for the baseline Faster RCNN (Fig.~\ref{fig:single}), ‘W/O Group’ indicates the naive multi-grained branch without grouping (Fig.~\ref{fig:twostage}), ‘W Group’ refers to the multi-grained branch with superclass grouping (Fig.~\ref{fig:twostage}), ‘Merged’ represents the merged heads approach (Fig.~\ref{fig:mergehead}).}
\label{tab:oiv4clothingresults}
\end{table*}

As we can see, it is beneficial to use information from different abstraction levels (in contrast to just the top one) to produce final detection results. The multi-grained branch with superclass grouping can clearly improve mAP scores by up to 2\%. In general, the multi-grained branch utilizing superclass information is better than other multi-grained solutions on the two datasets. This shows that the injected superclass knowledge (at nearly no cost) helps with overall detection. It is worth mentioning that despite a lower overall mAP score, the merged head actually achieves comparable mAP50 scores (slightly lower for DeepFashion2 and slightly higher for OpenImagesV4-Clothing). A possible contributing factor here is the relatively large amount of quality training data, especially in the larger OpenImagesV4-Clothing dataset. With enough training data, the network can possibly learn some hierarchical information itself. For this architecture to work well, much more data and annotations are needed. Comparing the first two rows in Table~\ref{tab:deepfashion2results} and~\ref{tab:oiv4clothingresults}, we can see that the naive multi-grained branches without category grouping does not help much over the baseline. In some cases, the no-grouping solution even leads to worse performance than the baseline. The possible reason is the confusion that the W/O Group method introduces by injecting the same finer grained category information in learning different-level features for both branches.
%more discussion on cases of small scale objects, possible reason for performance degradation in the DF2 case

\subsubsection{Per-category AP}
In addition to overall mAP, we report per-category AP in this section. The category columns are sorted in ascending order of the number of instances. It is evident that both datasets are imbalanced. In DeepFashion2, the largest category has 71,645 instances while the smallest category has only 543 instances (130x smaller). In OpenImagesV4-Clothing, there are 500x more samples in the largest category than in the smallest category. A good overall detection score may be misleading if the small classes are what we care more about. Table~\ref{tab:percategorydf2} and Table~\ref{tab:percategoryoiv4} demonstrate the detailed per-category AP scores for all four frameworks.
\begin{table*}[!ht]
\begin{center}
\begin{tabular}{|c|c|c|c|c|c|c|}
\hline
%& S. S. outwear & Sling & Sling Dress & L. S. dress & L. S. outwear & Vest \\
& Short sleeve outwear & Sling & Sling dress & Long sleeve dress & Long sleeve outwear & Vest \\
\hline
Num & 543 & 1985 & 6492 & 7907 & 13457 & 16095 \\
\hline
Baseline & 42.45 & 47.55 & 65.40 & 53.43 & 72.57 & 66.93 \\
W/O Grp & 44.29 & 48.12 & 66.74 & 53.21 & 73.80 & 67.47 \\
Merged & 46.11 & 49.63 & 68.41 & 56.28 & 74.22 & 68.68 \\
W Group & 46.51 & 51.59 & 68.24 & 57.06 & 73.97 & 68.28 \\
\hline
\end{tabular}
\vskip 0.1in
\begin{tabular}{|c|c|c|c|c|c|c|c|}
\hline
& Short sleeve dress & Vest dress & Skirt & Long sleeve shirt & Shorts & Trousers & Short sleeve shirt \\
\hline
Num & 17211 & 17949 & 30835 & 36064 & 36616 & 55387 & 71645 \\
\hline
Baseline & 72.31 & 72.69 & 74.20 & 73.26 & 71.08 & 75.32 & 81.12 \\
W/O Grp & 72.24 & 72.95 & 74.78 & 74.19 & 72.24 & 76.20 & 81.82 \\
Merged & 74.01 & 74.83 & 76.18 & 74.54 & 72.93 & 76.40 & 82.18 \\
W Group & 73.73 & 74.81 & 76.31 & 74.67 & 73.46 & 76.64 & 82.65 \\
\hline
\end{tabular}
\end{center}
\caption{Per-category AP of the DeepFashion2. The columns are sorted in ascending order of the number of instances in that category. %S.S.: short sleeve, L.S.: long sleeve. 
\textbf{Num} refers to the number of training bounding box instances in that category. ‘Baseline’: Fig.~\ref{fig:single}, ‘W/O Grp’: Fig.~\ref{fig:twostage} without grouping, ‘W Group’: Fig.~\ref{fig:twostage} with grouping, ‘Merged’: Fig.~\ref{fig:mergehead}.}
\label{tab:percategorydf2}
\end{table*}

\begin{table*}[!ht]
\begin{center}
\begin{tabular}{|c|c|c|c|c|c|c|c|c|c|c|}
\hline
& \shortstack{Brief- \\ case} & Tiara & Swimcap & Suitcase & Sombrero & Crown & Miniskirt & Backpack & Sock & Scarf \\
\hline
Num & 155 & 411 & 591 & 602 & 644 & 684 & 954 & 1205 & 1259 & 2280 \\
\hline
Baseline & 4.81 & 51.20 & 49.16 & 33.90 & 38.04 & 47.55 & 51.19 & 43.90 & 57.82 & 37.74 \\
W/O Grp & 9.79 & 47.66 & 45.03 & 30.21 & 29.36 & 49.67 & 57.90 & 45.61 & 61.32 & 38.76 \\
Merged & 12.28 & 47.05 & 47.41 & 35.05 & 42.89 & 50.14 & 57.90 & 48.00 & 63.25 & 40.85 \\
W Group & 12.98 & 55.40 & 47.99 & 36.10 & 48.00 & 51.73 & 57.87 & 45.62 & 63.19 & 41.09 \\
\hline
\hline
& \shortstack{Hand-\\bag} & Necklace & Sandal & \shortstack{Cowboy\\ hat} & \shortstack{High\\heels} & Boot & Fedora & \shortstack{Roller\\skates} & Coat & Sunhat \\
\hline
Num & 2425 & 2694 & 2912 & 3034 & 3093 & 3113 & 3624 & 5283 & 6434 & 6935 \\
\hline
Baseline & 76.53 & 60.50 & 40.28 & 38.61 & 42.52 & 57.86 & 58.55 & 44.18 & 43.77 & 50.14 \\
W/O Grp & 78.32 & 62.21 & 40.97 & 41.22 & 42.18 & 59.27 & 64.64 & 45.49 & 43.73 & 52.69 \\
Merged & 78.49 & 63.86 & 43.26 & 35.63 & 40.09 & 58.16 & 66.11 & 44.15 & 46.16 & 51.25 \\
W Group & 78.76 & 62.10 & 44.34 & 41.99 & 42.47 & 58.03 & 62.79 & 44.58 & 44.76 & 49.63 \\
\hline
\hline
& Shirt & Goggles & Tie & \shortstack{Football \\ helmet} & \shortstack{Bike\\helmet} & Shorts & Sunglasses & Jacket & Glasses & Jeans \\
\hline
Num & 7416 & 9593 & 10535 & 11657 & 15927 & 16971 & 23956 & 25782 & 57857 & 78348 \\
\hline
Baseline & 59.49 & 37.36 & 72.11 & 52.12 & 41.31 & 41.74 & 31.51 & 41.92 & 44.05 & 56.33 \\
W/O Grp & 63.05 & 35.85 & 72.94 & 53.38 & 44.55 & 41.28 & 33.60 & 41.81 & 46.81 & 58.43 \\
Merged & 64.32 & 37.55 & 73.59 & 53.46 & 43.86 & 42.87 & 31.97 & 42.96 & 46.90 & 58.83 \\
W Group & 64.34 & 38.28 & 72.95 & 52.48 & 44.03 & 40.59 & 33.06 & 41.73 & 46.12 & 57.95 \\
\hline
\end{tabular}
\end{center}
\caption{Per-category AP of the OpenImagesV4-Clothing. Only grouped categories are shown. The columns are sorted in ascending order of the number of instances. Num refers to the number of training bounding box instances in that category. ‘Baseline’: Fig.~\ref{fig:single}, ‘W/O Grp’: Fig.~\ref{fig:twostage} without grouping, ‘W Group’: Fig.~\ref{fig:twostage} with grouping, ‘Merged’: Fig.~\ref{fig:mergehead}.}
\label{tab:percategoryoiv4}
\end{table*}

\begin{figure*}
	\centering
	\includegraphics[height=2.85in]{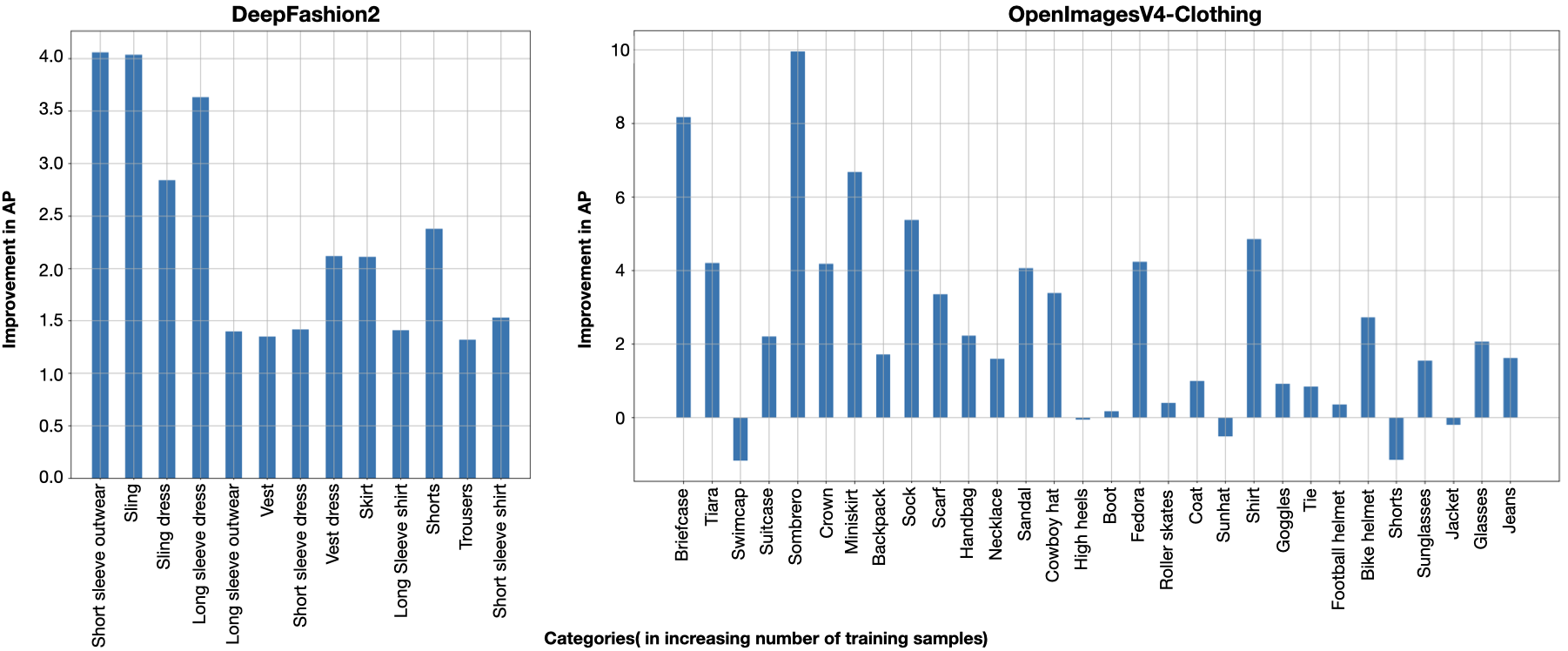}
	\caption{Improvement on AP for DeepFashion2 (left) and OpenImagesV4-Clothing (right) datsets. The categories are sorted in ascending order of number of training samples. We can see that the improvements are significant for categories with a small number of samples.}
	\label{fig:improvement_map}
\end{figure*}

We can see from tables~\ref{tab:percategorydf2} and \ref{tab:percategoryoiv4} that, in general, injecting superclass grouping information to the coarse branch helps with small classes more than larger classes. On average, we improve the overall mAP over the baseline method by 2.3 and 2.5 for the DeepFashion2 and OpenImagesV4-Clothing datasets, respectively. More importantly, we improve the mAP for last 30\% categories (in terms of number of training samples) by 2.55 and 4.59 for the two datasets, respectively. In particular, the mAP score is improved by as high as about 9.96 for the small \emph{sombrero} category in OpenImages-Clothing that contains only 644 instances. To better illustrate the general trend that small classes benefit more, we present the improvement in AP for all categories and for both datasets in Figure~\ref{fig:improvement_map}. 
One possible reason for the trend is that similar classes can share much low level features. By explicitly grouping them together, it helps to learn such features discriminatively, which is especially helpful for minority classes. In addition, category grouping may alleviate the over-fitting issue when attempting to train millions of parameters using only a few hundred instances for a class. That said, there are some small category exceptions where the performance improvement is small or even negative. This could be caused by the imperfection of superclass grouping. Also, different categories have different difficulties to be detected/recognized. Information from closely-related categories may not be enough to greatly improve the performance for some challenging categories.

\subsubsection{Qualitative results}
\begin{figure*}
%\begin{subfigure}{0.195\linewidth}
%    \begin{center}
%    \includegraphics[height=1.55in]{img_orig.png}
%%    \caption*{Original Image}
%    \end{center}
%\end{subfigure}
%\begin{subfigure}{0.195\linewidth}
%    \begin{center}
%    \includegraphics[height=1.55in]{img_baseline.png}
%%    \caption*{Baseline}
%    \end{center}
%\end{subfigure}
%\begin{subfigure}{0.195\linewidth}
%    \begin{center}
%    \includegraphics[height=1.55in]{img_without_grouping.png}
%%    \caption*{W/O grouping}
%    \end{center}
%\end{subfigure}
%\begin{subfigure}{0.195\linewidth}
%    \begin{center}
%    \includegraphics[height=1.55in]{img_merged.png}
%%    \caption*{Merged}
%    \end{center}
%\end{subfigure}
%\begin{subfigure}{0.195\linewidth}
%    \begin{center}
%    \includegraphics[height=1.55in]{img_with_grouping.png}
%%    \caption*{With grouping}
%    \end{center}
%\end{subfigure}
%\\\vskip 0.1in
\begin{subfigure}{0.195\linewidth}
    \begin{center}
    \includegraphics[height=1.55in, trim=4cm 3.5cm 3cm 4.5cm, clip]{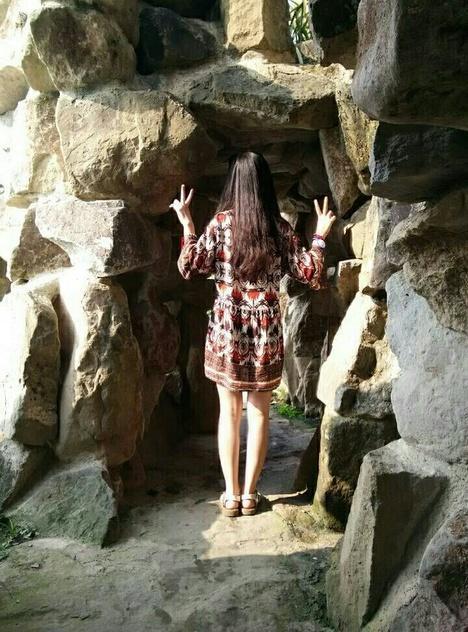}
%    \caption*{Original Image}
    \end{center}
\end{subfigure}
\begin{subfigure}{0.195\linewidth}
    \begin{center}
    \includegraphics[height=1.55in, trim=4cm 3.5cm 3cm 4.5cm, clip]{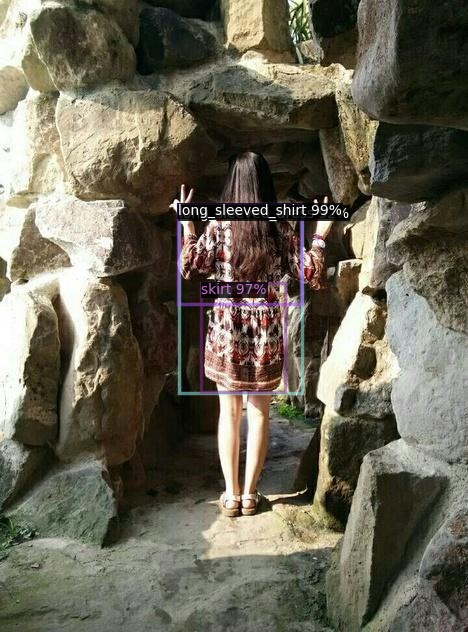}
%    \caption*{Baseline}
    \end{center}
\end{subfigure}
\begin{subfigure}{0.195\linewidth}
    \begin{center}
    \includegraphics[height=1.55in, trim=4cm 3.5cm 3cm 4.5cm, clip]{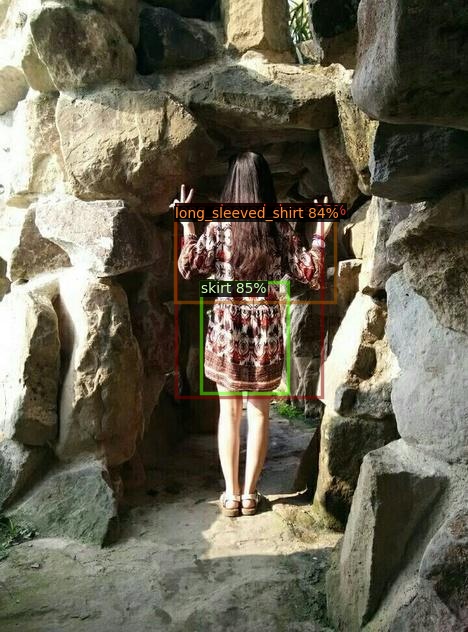}
%    \caption*{W/O grouping}
    \end{center}
\end{subfigure}
\begin{subfigure}{0.195\linewidth}
    \begin{center}
    \includegraphics[height=1.55in, trim=4cm 3.5cm 3cm 4.5cm, clip]{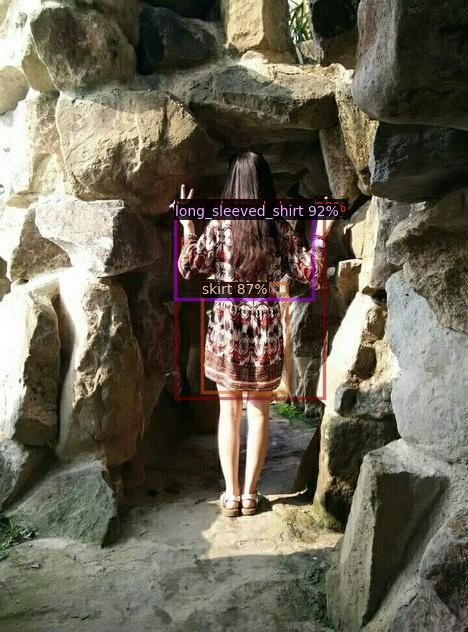}
%    \caption*{Merged}
    \end{center}
\end{subfigure}
\begin{subfigure}{0.195\linewidth}
    \begin{center}
    \includegraphics[height=1.55in, trim=4cm 3.5cm 3cm 4.5cm, clip]{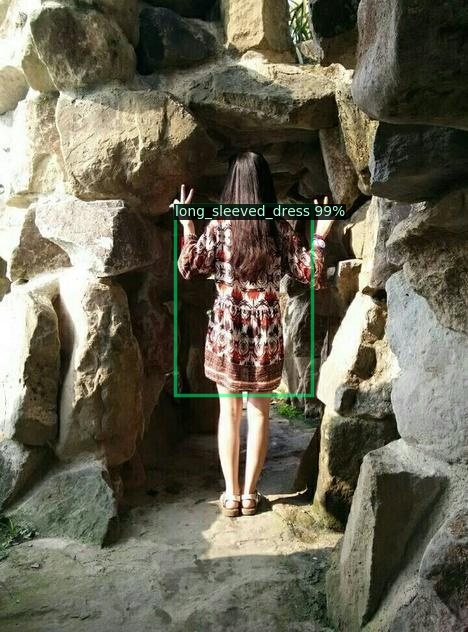}
%    \caption*{With grouping}
    \end{center}
\end{subfigure}
\\\vskip 0.1in
\begin{subfigure}{0.195\linewidth}
    \begin{center}
    \includegraphics[height=1.55in]{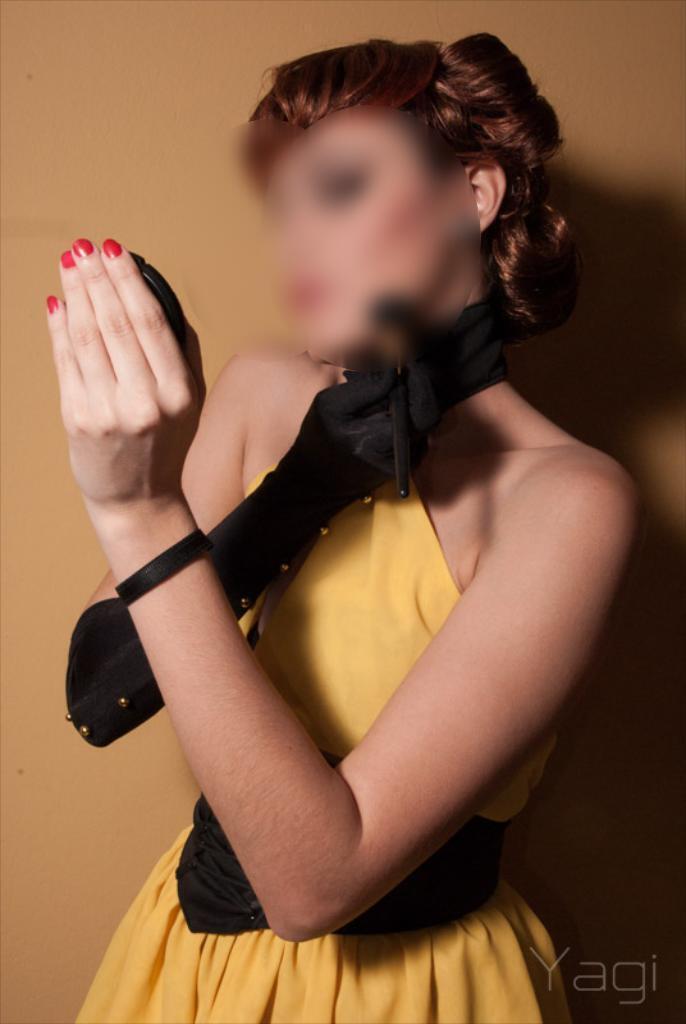}
    \caption*{Original Image}
    \end{center}
\end{subfigure}
\begin{subfigure}{0.195\linewidth}
    \begin{center}
    \includegraphics[height=1.55in]{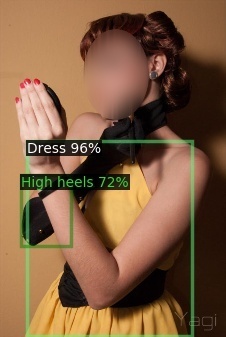}
    \caption*{Baseline}
    \end{center}
\end{subfigure}
\begin{subfigure}{0.195\linewidth}
    \begin{center}
    \includegraphics[height=1.55in]{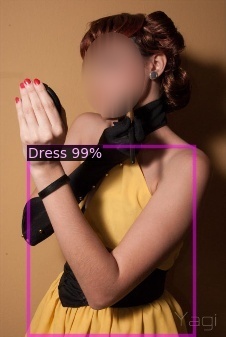}
    \caption*{W/O grouping}
    \end{center}
\end{subfigure}
\begin{subfigure}{0.195\linewidth}
    \begin{center}
    \includegraphics[height=1.55in]{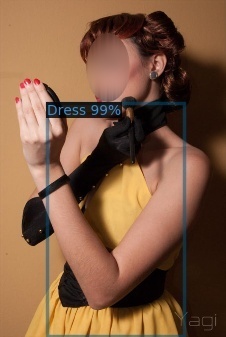}
    \caption*{Merged}
    \end{center}
\end{subfigure}
\begin{subfigure}{0.195\linewidth}
    \begin{center}
    \includegraphics[height=1.55in]{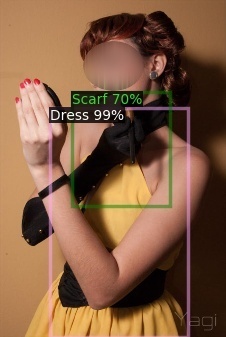}
    \caption*{With grouping}
    \end{center}
\end{subfigure}
\caption{\textbf{Best viewed in color.}Qualitative results of different heads with single and multi-grained branches. W/O grouping: naive multi-grained branches without grouping, Merged: multi-depth branches with merged heads, W Grouping: multi-grained branches with superclass grouping. The first image is from DeepFashion2 (cropped for clarity) and the second image is from OpenImagesV4-Clothing. We have blurred faces for privacy reasons.}
\label{fig:qualitativeresults}
\end{figure*}

\begin{figure*}
	\centering
	\includegraphics[width=0.95\linewidth]{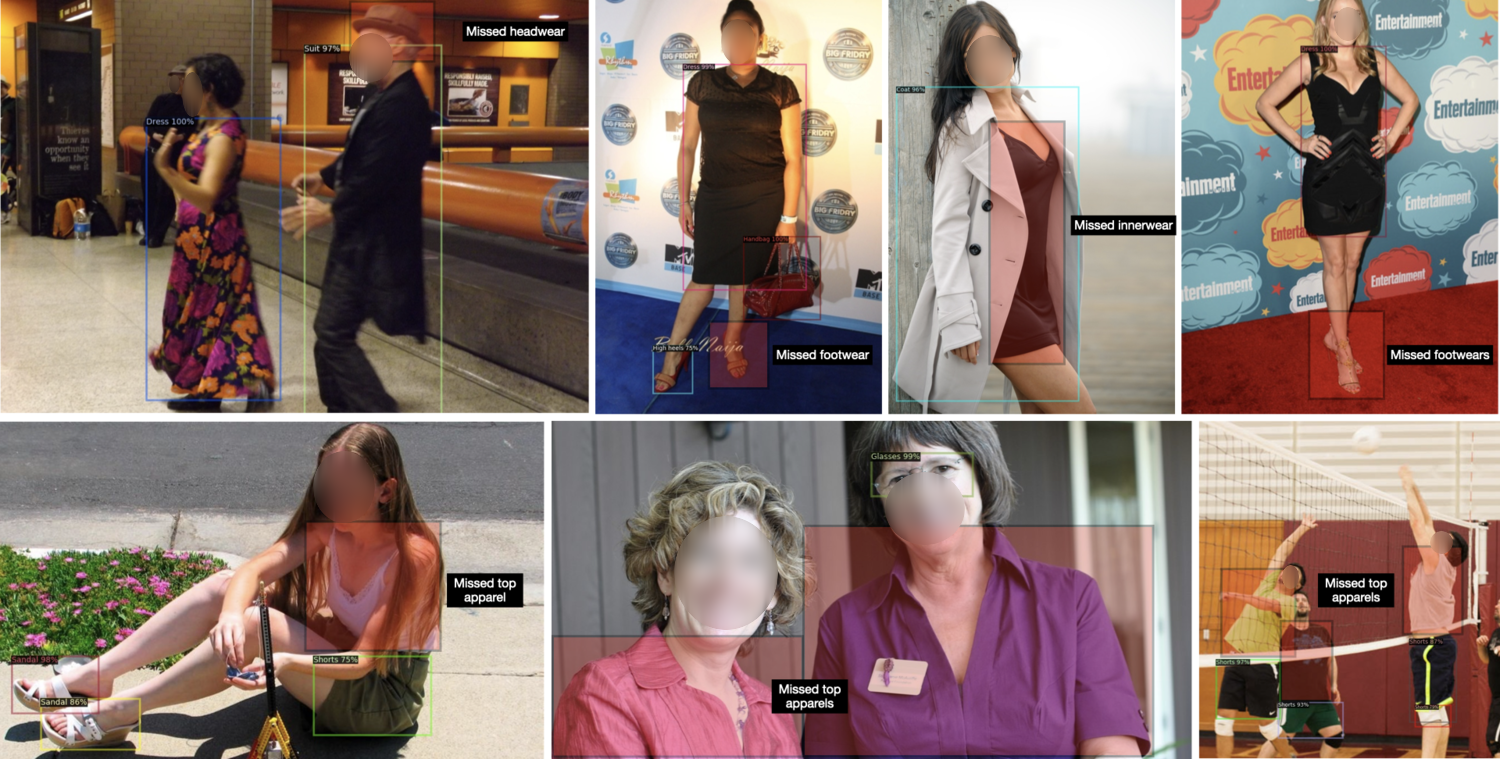}	
	\caption{\textbf{Best viewed in color.} Examples of failure cases of our proposed solution on OpenImagesV4-Clothing dataset. We highlight the missed detections with red boxes and also specify what category we missed to detect. We have blurred faces for privacy reasons.}
	\label{fig:failure_case}
\end{figure*}

Figure~\ref{fig:qualitativeresults} demonstrates some successful qualitative examples on the two datasets. These results give an idea how multi-grained branches with superclass grouping can improve over the baseline.
In Figure~\ref{fig:failure_case}, we depict some of our typical failure cases. As we can see, our failure cases correspond mostly to false negatives. %, \emph{i.e.,}, we sometimes miss to detect certain apparel categories.
%-------------------------------------------------------------------------
\section{Discussion and future work}\label{sec:futurework}

Different from popular deep layer aggregation strategies \cite{yu2018deep}, we have attempted to use separate branches to capture coarse and fine features. The reason is that although human hierarchy knowledge is useful (especially when labeled data is limited), it may not be well-aligned with the coarse-to-fine features that is deeply learned using a single branch with only fine-grained supervision on top. That said, through shared lower layer features, the separate branches can interact with each other. This perhaps explains why our multibranch solution (Fig.~\ref{fig:twostage}) can perform better than the merged solution (Fig.~\ref{fig:mergehead}) that does not intake any human knowledge about the superclass.
Apart from human hierarchy knowledge, other superclass grouping strategies can be explored or even learned if we have enough labeled data. Although both datasets we used are already very large publicly available datasets in this field, it would be of interest to experiment on even larger data and more annotations. It is possible that with unlimited data and annotations, merged head can freely learn to integrate different abstractness level information to minimize the loss. That said, in most real-world cases with limited annotated data, superclass grouping as shown can be a simple yet effective way to improve detection performance.

In addition to improving detection performance, extra levels of supervision offer more flexibility to utilize information at different granularities. When two or more datasets have multi-grained groundtruth labels, it is hard to put them on a flat level (\textit{e.g.}, dog and golden retriever). In such cases, multi-grained branches are promising to integrate information from these datasets. Also, our frameworks with multi-grained branches may handle newer categories more easily. If the new category's superclass already exists in our model, the retraining efforts can be focused more on the fine-grained branch than the coarse-grained branch and the lower layers. Such explorations are deferred to future work.

%------------------------------------------------------------------------
\section{Conclusion}\label{sec:conclusion}

In this paper, we study several detection head solutions with multi-grained branches for Faster RCNN backends which can make use of ROI information from different abstractness levels. According to our experiments on DeepFashion2 and OpenImagesV4-Clothing datasets, detection heads with multi-grained branches can boost detection performance by up to 2\% without requiring extra expensive annotations. Particularly, superclass grouping with human knowledge can greatly improve the performance for categories with fewer images. For example, in the sombrero case of OpenImagesV4-Clothing, its mAP is increased by as high as 10\%.

%\clearpage
{\small
\bibliographystyle{ieee_fullname}
\bibliography{egbib}

\begin{thebibliography}{10}\itemsep=-1pt

\bibitem{cai2018cascade}
Zhaowei Cai and Nuno Vasconcelos.
\newblock Cascade r-cnn: Delving into high quality object detection.
\newblock In {\em Proceedings of the IEEE conference on computer vision and
  pattern recognition}, pages 6154--6162, 2018.

\bibitem{chawla2002smote}
Nitesh~V. Chawla, Kevin~W. Bowyer, Lawrence~O. Hall, and W.~Philip Kegelmeyer.
\newblock Smote: synthetic minority over-sampling technique.
\newblock {\em Journal of artificial intelligence research}, 16:321--357, 2002.

\bibitem{imagenet_cvpr09}
J. Deng, W. Dong, R. Socher, L.-J. Li, K. Li, and L. Fei-Fei.
\newblock {ImageNet: A Large-Scale Hierarchical Image Database}.
\newblock In {\em CVPR09}, 2009.

\bibitem{deng2012hedging}
Jia Deng, Jonathan Krause, Alexander~C Berg, and Li Fei-Fei.
\newblock Hedging your bets: Optimizing accuracy-specificity trade-offs in
  large scale visual recognition.
\newblock In {\em 2012 IEEE Conference on Computer Vision and Pattern
  Recognition}, pages 3450--3457. IEEE, 2012.

\bibitem{ge2019deepfashion2}
Yuying Ge, Ruimao Zhang, Xiaogang Wang, Xiaoou Tang, and Ping Luo.
\newblock Deepfashion2: A versatile benchmark for detection, pose estimation,
  segmentation and re-identification of clothing images.
\newblock In {\em Proceedings of the IEEE Conference on Computer Vision and
  Pattern Recognition}, pages 5337--5345, 2019.

\bibitem{gidaris2015object}
Spyros Gidaris and Nikos Komodakis.
\newblock Object detection via a multi-region and semantic segmentation-aware
  cnn model.
\newblock In {\em Proceedings of the IEEE international conference on computer
  vision}, pages 1134--1142, 2015.

\bibitem{gidaris2016attend}
Spyros Gidaris and Nikos Komodakis.
\newblock Attend refine repeat: Active box proposal generation via in-out
  localization.
\newblock {\em arXiv preprint arXiv:1606.04446}, 2016.

\bibitem{girshick2015fast}
Ross Girshick.
\newblock Fast r-cnn.
\newblock In {\em Proceedings of the IEEE international conference on computer
  vision}, pages 1440--1448, 2015.

\bibitem{girshick2014rich}
Ross Girshick, Jeff Donahue, Trevor Darrell, and Jitendra Malik.
\newblock Rich feature hierarchies for accurate object detection and semantic
  segmentation.
\newblock In {\em Proceedings of the IEEE conference on computer vision and
  pattern recognition}, pages 580--587, 2014.

\bibitem{he2009learning}
Haibo He and Edwardo~A Garcia.
\newblock Learning from imbalanced data.
\newblock {\em IEEE Transactions on knowledge and data engineering},
  21(9):1263--1284, 2009.

\bibitem{he2017mask}
Kaiming He, Georgia Gkioxari, Piotr Doll{\'a}r, and Ross Girshick.
\newblock Mask r-cnn.
\newblock In {\em Proceedings of the IEEE international conference on computer
  vision}, pages 2961--2969, 2017.

\bibitem{he2016deep}
Kaiming He, Xiangyu Zhang, Shaoqing Ren, and Jian Sun.
\newblock Deep residual learning for image recognition.
\newblock In {\em Proceedings of the IEEE conference on computer vision and
  pattern recognition}, pages 770--778, 2016.

\bibitem{huang2017densely}
Gao Huang, Zhuang Liu, Laurens Van Der~Maaten, and Kilian~Q Weinberger.
\newblock Densely connected convolutional networks.
\newblock In {\em Proceedings of the IEEE conference on computer vision and
  pattern recognition}, pages 4700--4708, 2017.

\bibitem{khan2017cost}
Salman~H Khan, Munawar Hayat, Mohammed Bennamoun, Ferdous~A Sohel, and Roberto
  Togneri.
\newblock Cost-sensitive learning of deep feature representations from
  imbalanced data.
\newblock {\em IEEE transactions on neural networks and learning systems},
  29(8):3573--3587, 2017.

\bibitem{krawczyk2014cost}
Bartosz Krawczyk, Micha{\l} Wo{\'z}niak, and Gerald Schaefer.
\newblock Cost-sensitive decision tree ensembles for effective imbalanced
  classification.
\newblock {\em Applied Soft Computing}, 14:554--562, 2014.

\bibitem{kuznetsova2018open}
Alina Kuznetsova, Hassan Rom, Neil Alldrin, Jasper Uijlings, Ivan Krasin, Jordi
  Pont-Tuset, Shahab Kamali, Stefan Popov, Matteo Malloci, Tom Duerig, et~al.
\newblock The open images dataset v4: Unified image classification, object
  detection, and visual relationship detection at scale.
\newblock {\em arXiv preprint arXiv:1811.00982}, 2018.

\bibitem{lin2017feature}
Tsung-Yi Lin, Piotr Doll{\'a}r, Ross Girshick, Kaiming He, Bharath Hariharan,
  and Serge Belongie.
\newblock Feature pyramid networks for object detection.
\newblock In {\em Proceedings of the IEEE conference on computer vision and
  pattern recognition}, pages 2117--2125, 2017.

\bibitem{lin2017focal}
Tsung-Yi Lin, Priya Goyal, Ross Girshick, Kaiming He, and Piotr Doll{\'a}r.
\newblock Focal loss for dense object detection.
\newblock In {\em Proceedings of the IEEE international conference on computer
  vision}, pages 2980--2988, 2017.

\bibitem{lin2014microsoft}
Tsung-Yi Lin, Michael Maire, Serge Belongie, James Hays, Pietro Perona, Deva
  Ramanan, Piotr Doll{\'a}r, and C~Lawrence Zitnick.
\newblock Microsoft coco: Common objects in context.
\newblock In {\em European conference on computer vision}, pages 740--755.
  Springer, 2014.

\bibitem{liu2016ssd}
Wei Liu, Dragomir Anguelov, Dumitru Erhan, Christian Szegedy, Scott Reed,
  Cheng-Yang Fu, and Alexander~C Berg.
\newblock Ssd: Single shot multibox detector.
\newblock In {\em European conference on computer vision}, pages 21--37.
  Springer, 2016.

\bibitem{miller1990introduction}
George~A Miller, Richard Beckwith, Christiane Fellbaum, Derek Gross, and
  Katherine~J Miller.
\newblock Introduction to wordnet: An on-line lexical database.
\newblock {\em International journal of lexicography}, 3(4):235--244, 1990.

\bibitem{redmon2016you}
Joseph Redmon, Santosh Divvala, Ross Girshick, and Ali Farhadi.
\newblock You only look once: Unified, real-time object detection.
\newblock In {\em Proceedings of the IEEE conference on computer vision and
  pattern recognition}, pages 779--788, 2016.

\bibitem{redmon2017yolo9000}
Joseph Redmon and Ali Farhadi.
\newblock Yolo9000: better, faster, stronger.
\newblock In {\em Proceedings of the IEEE conference on computer vision and
  pattern recognition}, pages 7263--7271, 2017.

\bibitem{redmon2018yolov3}
Joseph Redmon and Ali Farhadi.
\newblock Yolov3: An incremental improvement.
\newblock {\em arXiv preprint arXiv:1804.02767}, 2018.

\bibitem{ren2015faster}
Shaoqing Ren, Kaiming He, Ross Girshick, and Jian Sun.
\newblock Faster r-cnn: Towards real-time object detection with region proposal
  networks.
\newblock In {\em Advances in neural information processing systems}, pages
  91--99, 2015.

\bibitem{sermanet2013overfeat}
Pierre Sermanet, David Eigen, Xiang Zhang, Micha{\"e}l Mathieu, Rob Fergus, and
  Yann LeCun.
\newblock Overfeat: Integrated recognition, localization and detection using
  convolutional networks.
\newblock {\em arXiv preprint arXiv:1312.6229}, 2013.

\bibitem{ting2000comparative}
Kai~Ming Ting.
\newblock A comparative study of cost-sensitive boosting algorithms.
\newblock In {\em In Proceedings of the 17th International Conference on
  Machine Learning}. Citeseer, 2000.

\bibitem{uijlings2013selective}
Jasper~RR Uijlings, Koen~EA Van De~Sande, Theo Gevers, and Arnold~WM Smeulders.
\newblock Selective search for object recognition.
\newblock {\em International journal of computer vision}, 104(2):154--171,
  2013.

\bibitem{wang2016training}
Shoujin Wang, Wei Liu, Jia Wu, Longbing Cao, Qinxue Meng, and Paul~J Kennedy.
\newblock Training deep neural networks on imbalanced data sets.
\newblock In {\em 2016 international joint conference on neural networks
  (IJCNN)}, pages 4368--4374. IEEE, 2016.

\bibitem{wang2017learning}
Yu-Xiong Wang, Deva Ramanan, and Martial Hebert.
\newblock Learning to model the tail.
\newblock In {\em Advances in Neural Information Processing Systems}, pages
  7029--7039, 2017.

\bibitem{wu2019detectron2}
Yuxin Wu, Alexander Kirillov, Francisco Massa, Wan-Yen Lo, and Ross Girshick.
\newblock Detectron2.
\newblock \url{https://github.com/facebookresearch/detectron2}, 2019.

\bibitem{yang2016craft}
Bin Yang, Junjie Yan, Zhen Lei, and Stan~Z Li.
\newblock Craft objects from images.
\newblock In {\em Proceedings of the IEEE Conference on Computer Vision and
  Pattern Recognition}, pages 6043--6051, 2016.

\bibitem{yang2019detecting}
Hao Yang, Hao Wu, and Hao Chen.
\newblock Detecting 11k classes: Large scale object detection without
  fine-grained bounding boxes.
\newblock In {\em Proceedings of the IEEE International Conference on Computer
  Vision}, pages 9805--9813, 2019.

\bibitem{yu2018deep}
Fisher Yu, Dequan Wang, Evan Shelhamer, and Trevor Darrell.
\newblock Deep layer aggregation.
\newblock In {\em Proceedings of the IEEE conference on computer vision and
  pattern recognition}, pages 2403--2412, 2018.

\bibitem{zhou2005training}
Zhi-Hua Zhou and Xu-Ying Liu.
\newblock Training cost-sensitive neural networks with methods addressing the
  class imbalance problem.
\newblock {\em IEEE Transactions on knowledge and data engineering},
  18(1):63--77, 2005.

\bibitem{zhou2010multi}
Zhi-Hua Zhou and Xu-Ying Liu.
\newblock On multi-class cost-sensitive learning.
\newblock {\em Computational Intelligence}, 26(3):232--257, 2010.

\bibitem{zitnick2014edge}
C~Lawrence Zitnick and Piotr Doll{\'a}r.
\newblock Edge boxes: Locating object proposals from edges.
\newblock In {\em European conference on computer vision}, pages 391--405.
  Springer, 2014.

\end{thebibliography}
}

\end{document}